\pdfoutput=1

\documentclass[11pt,table,xcdraw]{article}
\usepackage{colortbl}
\usepackage{svg}
\usepackage{subfigure}
\usepackage{paralist, tabularx}
\newcommand\blfootnote[1]{%
  \begingroup
  \renewcommand\thefootnote{}\footnote{#1}%
  \addtocounter{footnote}{-1}%
  \endgroup
}
\usepackage{soul}
\usepackage{array}
\usepackage{multirow} 
\usepackage{makecell}
\usepackage{acl}

\usepackage{times}
\usepackage{latexsym}

\usepackage[T1]{fontenc}

\usepackage[utf8]{inputenc}

\usepackage{microtype}

\title{Goal-Directed Story Generation: Augmenting Generative Language Models with Reinforcement Learning}

\author{Amal Alabdulkarim $^{\dagger *}$, Winston Li$^{\dagger *}$, Lara J. Martin $ ^{\ddagger}$, Mark O. Riedl $ ^{\dagger}$
        \\$ ^{\dagger}$ School of Interactive Computing, Georgia Institute of Technology
        \\$ ^{\ddagger}$ Computer and Information Science, University of Pennsylvania
  \\\texttt{\{amal,wli384,riedl\}@gatech.edu; laramar@seas.upenn.edu} }
\begin{document}
\maketitle
\blfootnote{* Denotes equal contribution.}

\begin{abstract}
The advent of large pre-trained generative language models has provided a common framework for AI story generation via sampling the model to create sequences that continue the story. 
However, sampling alone is insufficient for story generation.
In particular, it is hard to direct a language model to create stories to reach a specific goal event. 
We present two automated techniques grounded in deep reinforcement learning and reward shaping to control the plot of computer-generated stories. 
The first utilizes proximal policy optimization to fine-tune an existing transformer-based language model to generate text continuations but also be goal-seeking. 
The second extracts a knowledge graph from the unfolding story, which is used by a policy network with graph attention to select a candidate continuation generated by a language model.  
We report on automated metrics pertaining to how often stories achieve a given goal event as well as human participant rankings of coherence and overall story quality compared to baselines and ablations.
\end{abstract}

\section{Introduction}
\begin{figure}[t]
    \centering
    \includegraphics[width=\linewidth]{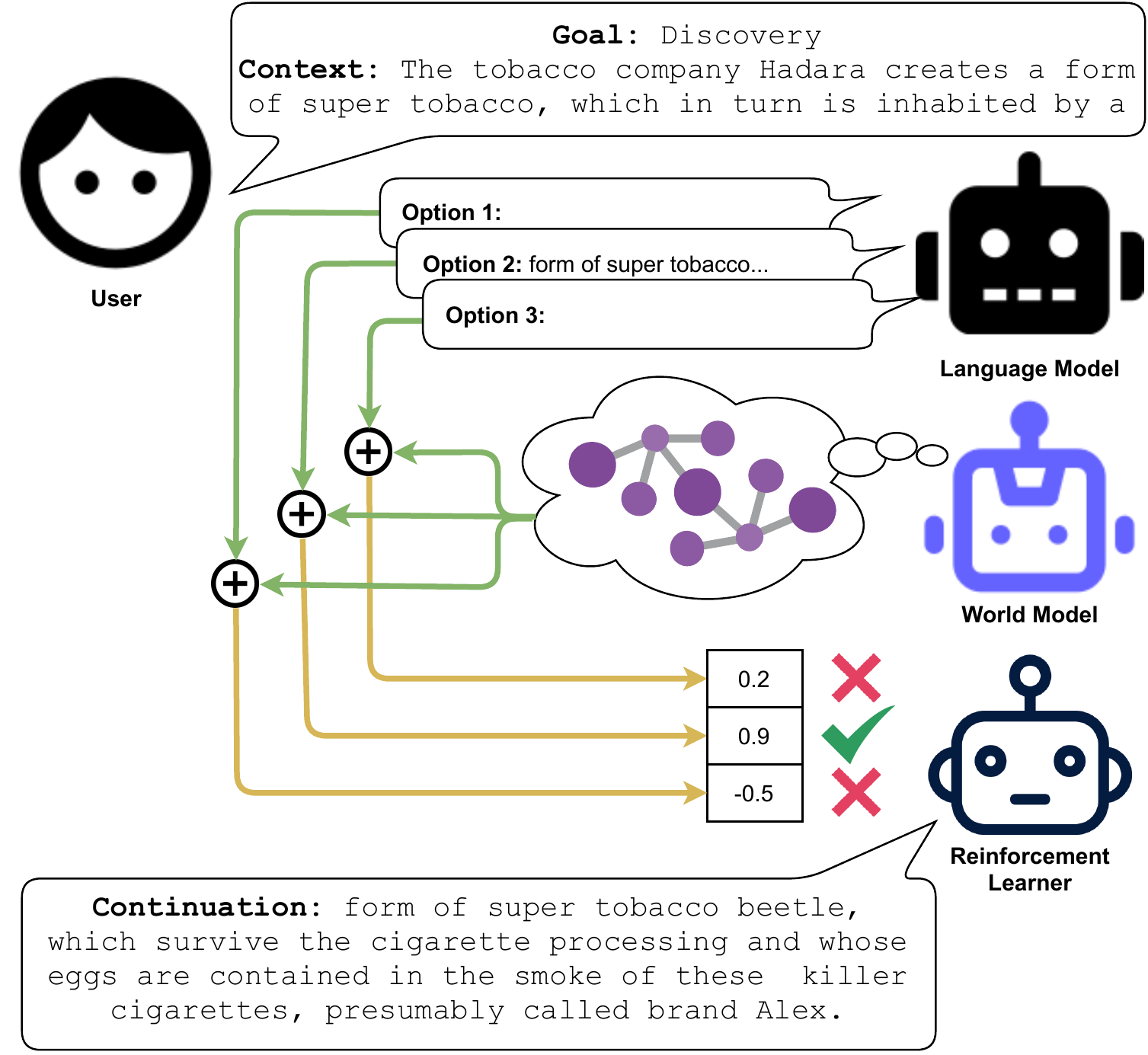}
    \caption{An example of a single iteration in our generation system.
    Given a goal and a context prompt, a language model is queried for a number of possible continuations.
    The system also maintains an abstract graph-based representation of the entities and relations in the story world.
    A reinforcement learning agent has learned to select options based on the abstract state and how likely the option is to move the story toward the given goal.
    Not shown: the world model is then updated and the continuation is added to the context for the next iteration.}
    \label{fig:example}
\end{figure}

Humans produce and consume stories every day in many different forms: news, entertainment, conversations, and more. Storytelling is crucial to the communication process, as humans engage with well-told stories and comprehend more information from stories \cite{suzuki2018dialogues}. Because of its effectiveness in human-to-human interactions, automating storytelling has become an important research focus. 

Automated Story Generation is the challenge of designing an artificial intelligence system that can generate a story from a minimal number of inputs---often just a prompt and some storytelling knowledge and/or storytelling model. 

Early work on story generation used planning~\cite{meehan1976metanovel,lebowitz1987planning,cavazza2003interacting,porteous2009controlling,riedl2010narrative,ware2010modeling} 
or case-based reasoning~\cite{perez2001mexica,peinado2005creativity,turner2014creative}.
In many cases, these systems are provided with a goal or outcome state.
For example a goal might be ``character X in jail'' or ``characters X and Y are married''.
However, these approaches require extensive domain knowledge engineering and rely on templated language.
Recently, large pre-trained neural language models have been applied to story generation because they circumvent the need for knowledge engineering and tend to produce relatively fluent, varied, and naturalistic language~\cite{roemmele2016writing,khalifa2017deeptingle,clark2018neural,martin2018event}.
To generate texts, language models are repeatedly invoked to generate the next token or sequence of tokens---called a {\em continuation}.
Language models are, however, not goal-directed. 
That is, one cannot naively provide both an context prompt and a goal to be achieved after an arbitrary number of continuations. 
Further, language models struggle with maintaining story coherence---the logical progression of events---and may also become repetitive.

Previous attempts to enhance the coherence of generated stories and control the trajectory of the story use conditioning on content-relevant features such as plot outlines~\cite{fan2018hierarchical,peng2018towards,rashkin2020plotmachines} or by hierarchical reasoning with abstract representations that help constrain story progression~\cite{martin2018event,yao2019plan,fan2019strategies,peng2021inferring}. 
These techniques do not address the need to make story generation systems goal-driven.

To make story generation systems goal-driven,
\citet{tambwekar2019controllable} 
trained a seq2seq language model on event abstractions (tuples containing verb, direct object, indirect object, and some other information) and then fine-tuned the model using reinforcement learning with a reward based on the average number of events it takes to get from one verb to a goal verb. 
They were able to show their models can achieve a given goal greater than 93\% of the time, and subjective ratings of story coherence improved significantly. However, the event abstractions are not human-readable without manual rewriting or a second event-to-sentence technique that often undoes any gains in coherence, as observed by an analysis of hierarchical story generation systems by \citet{ammanabrolu2020story}.

This paper has two aims.
First, we show that applying a reward shaping fine-tuning technique such as that by \citeauthor{tambwekar2019controllable} above does not directly translate to more modern large pre-trained language models such as GPT-2~\cite{radford2019language}.
Large pre-trained language models produce more natural language and can handle a larger range of inputs but, like seq2seq models, are not inherently goal-driven. 
Unfortunately, we observe that large language models are harder to control; our experiments with reward shaping based fine-tuning toward a given goal
only results in a 50\% goal achievement rate (although fluency of story outputs is greatly improved). 

Our second aim is to introduce a new technique in which a second {\em policy model} can be trained to guide a non-fine-tuned GPT-2 to a given end-goal, achieving 90+\% goal success rate while retaining the language model's fluency.
The key insight is that this second model operates on an abstracted state space represented as a knowledge graph---a set of $\langle subject, relation, object\rangle$ tuples.
This story world state representation explicitly captures the entities in a story and their relations instead of relying on the hidden state of the language model to accurately represent the state of the story world.
Given a knowledge graph representing the state of the story world, the policy model predicts the utility of the state, which is proportional to the number of sentences needed to achieve a goal.
Thus, GPT-2 generates plausible continuations while the policy model learns to select continuations based on how they move the story forward.

We report on a combination of automated and human participant evaluations.
We focus on evaluating our system in the domain of science fiction plots~\cite{ammanabrolu2020story}, consistent with prior work~\cite{tambwekar2019controllable,ammanabrolu2020story}.
Automated evaluation shows that our full two-model reinforcement learning technique achieves the desired-end goal $98.73\%$ of the time, which means that users can provide both a prompt and an ending.
Our human participant studies show that the full model is perceived to be more coherent than baseline alternatives.

\section{Background and Related Work}

In this section we situate our work in the area of neural approaches to story generation.\footnote{See \citet{gervas2009computational, kybartas_survey_2017} for surveys of non-neural approaches to story generation.}

Neural language model based approaches to story generation 
start with a given text prompt and generate story continuations by sampling from a learned distribution $P(tok_n | tok_{<n};\theta)$ where $\theta$ is the parameters that approximate the probability distribution from the underlying data---models trained on a corpus of story will produce text that appears to be a story~\cite{roemmele2016writing,khalifa2017deeptingle,clark2018neural,martin2018event}.
These techniques have improved with the adoption of large, pre-trained, transformer-based models, such as GPT-2~\cite{radford2019language}. 
These models can be fine-tuned on representative data from a particular domain.
However, larger models such as GPT-3~\cite{brown2020language}  may be behind closed APIs that do not allow fine-tuning, making it more important that we have solutions to the problem of controllable text generation that do not rely on fine-tuning of language models.

Neural story generators are inherently ``backwards-looking'' in the sense that they produce tokens or sequences that are likely to occur based on a window of prior tokens.
As a result, it is challenging to control the direction that the story will unfold and neural generated stories tend to meander or become repetitive.
One means of controlling story generation is to condition generation on high-level plot outlines~\cite{fan2018hierarchical,peng2018towards,rashkin2020plotmachines} or story in-filling~\cite{donahue2020ilm}.
However, these techniques assume a human or other source have already determined the key plot points and the generator provides missing details.
Coherence can also be increased by systems that generate their own plot-level abstractions and then condition a language model on those plot labels~\cite{yao2019plan,fan2019strategies,peng2021inferring}.
While shown to improve perceptions of narrative coherence, these techniques cannot guarantee a goal achievement.

One way to ensure goal achievement is to provide the final event/sentence of a story in addition to the first event/sentence of a story as inputs.
\citet{wang2020narrative} propose a generation-by-interpolation approach to story generation. 
Starting with the first and last sentence in a story, their method uses GPT-2 to generate several candidates to go in between and then chose based in its coherence with the first and last sentence.
The C2PO system~\cite{ammanabrolu20automated} uses bi-directional search from a given start and given end, operating in the space of character goals as inferred by the COMET commonsense inference model~\cite{bosselut2019comet}.
This system, however only generates short, templated sentences and can best be thought of as a high-level plot generator (Ammanabrolu et al. private correspondence).
The EDGAR system~\cite{louisedgar} generates backwards from a given last sentence but cannot guarantee a given first sentence.

As overviewed in the introduction,
\citet{tambwekar2019controllable} trained a LSTM-based sequence-to-sequence neural model to generate continuations while also increasing the likelihood of achieving a given goal event. 
The LSTM model produced {\em events}, which are tuples of the form $\langle s,v, o, m\rangle$ where $s$ is a subject, $v$ is a verb class, $o$ is a direct object, and $m$ is an additional piece of modifier information about the verb. 
While their model reliably achieved the given goal, sequences of event tuples are not human-readable, requiring either manual rewriting or a second model to translate events into human-readable sentences such as~\cite{ammanabrolu2020story}.
This pipeline approach is lossy in the sense that the translation from events to sentences can undo the coherence of event-to-event transitions.
Further, events generated may be out-of-distribution for the sentence generator.
This is especially prone to happen because the fine-tuning technique uses teacher forcing, which generates an event and the substitutes the verb for one known to be closer to the goal; this can result in unusual combinations of subjects, verbs, and objects. 
To determine whether a verb is closer to the goal, the system analyze the corpus and compute the average number of sentences between any two verbs.
They then cluster the verbs using Jenks Natural Breaks clustering~\cite{jenks1971error} and reward the system for generating events with verbs one cluster closer to the goal.
A natural progression to this work would be to apply the the reward-shaping based fine-tuning approach from \citeauthor{tambwekar2019controllable} to a large pre-trained language model.

Our proposed approach produces an abstraction of the story state in the form of a knowledge graph, a set of $\langle subject, relation, object\rangle$ tuples.
Unlike the events used by \citeauthor{tambwekar2019controllable}, the knowledge graph contains all known entities in the story world and any known relations between entities.

Knowledge graphs have been shown to improve natural language understanding in several domains, and much work has been done to generalize neural network models to graph structured data. 
Using a knowledge graph to augment a reinforcement learning agent has produced results in other domains.  For example, knowledge graphs were shown to improve agent decision-making in interactive narrative text game playing domains~\cite{kgdqnTextgames,ammanabrolu2020Graph,ammanabrolu2020avoid,ammanabrolu2021learning}. 

\section{Methodology}
\label{sec:methodology}

Our goal is to generate stories that (1) are able to reliably reach a specific goal, (2) follow a logical plot leading to the goal sentence, (3) and have reasonable lengths (ie. do not jump straight to generating a closing sentence). 
For our system a goal is any sentence that contains a verb that is a member of a specific, given VerbNet~\cite{schuler2005verbnet} class.
For example, the verbs: find, guess, and solve are members of the VerbNet class \texttt{discover-84}, which has thirteen verbs.

Our method depends on two models: a language model based on GPT-2~\cite{radford2019language} and a policy model trained via reinforcement learning to select alternative continuations that progress the story incrementally toward the goal.  
We will show that both models are needed to achieve all three of our objectives.

\subsection{Dataset and Preprocessing}
\label{prepdata}
We use the science fiction plot corpus~\cite{ammanabrolu2020story}. The dataset contains more than 1400 generalized science fiction stories of variable lengths. The stories have named entities replaced with tags denoting category and number within a story, maintaining consistency for named entities. We perform the following pre-processing steps:
\begin{enumerate}
\item \textbf{Dataset Splitting} We performed a 70:30 split on our training data, resulting in sets of size 1102 and 472 stories, respectively.
\item \textbf{Verb class extraction} For each sentence, we parse the sentences using SpaCy \cite{spacy}, and then we extract the verbs from the parsed sentences. We then lemmatize each verb and match it to its VerbNet class \cite{schuler2005verbnet}. 
\item \textbf{Tokenization} We use the Huggingface pre-trained GPT2 tokenizer
to tokenize our sentences and prepare our model's input queries.
\end{enumerate}
We fine-tune GPT-2 on the science fiction dataset; 
we refer to this model as {\em GPT2-sci-fi}. 

\begin{figure*}[t!]
    \centering
    \includegraphics[width=\textwidth]{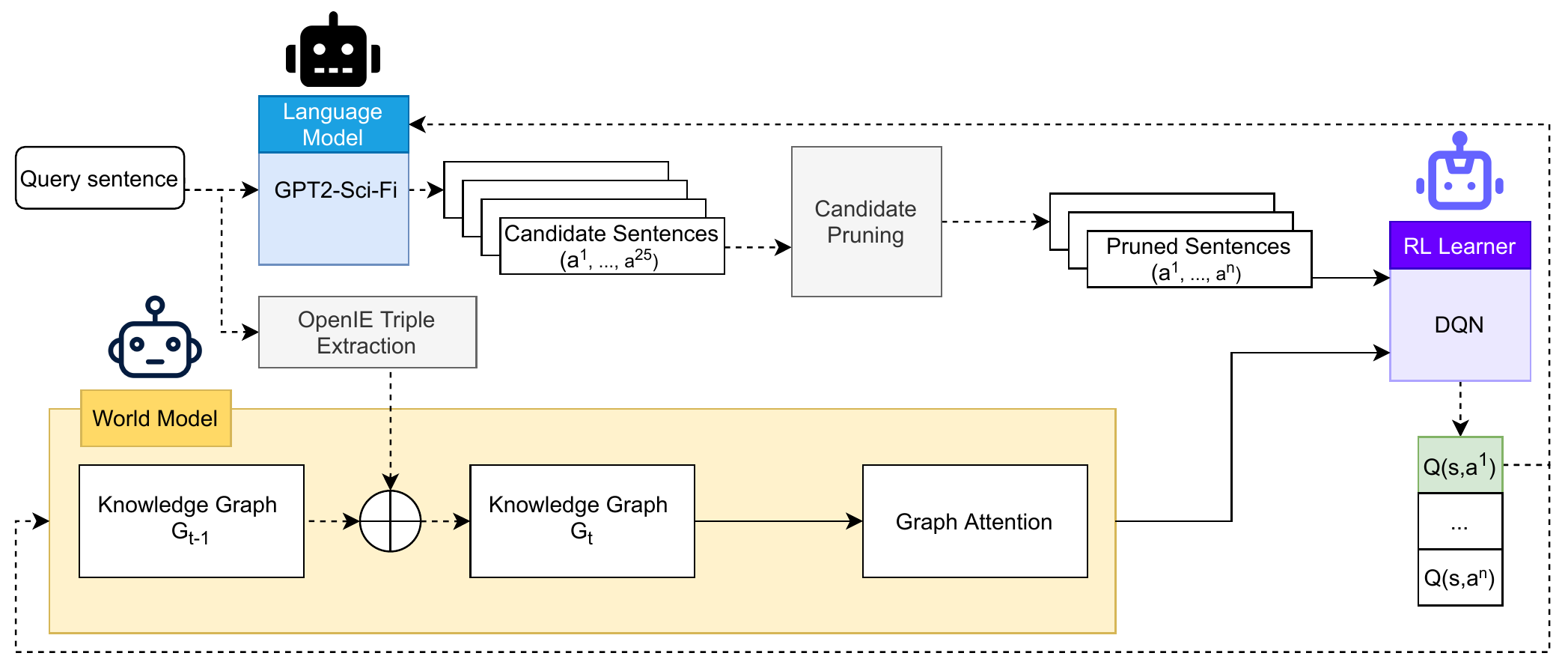}
    \caption{KG-DQN architecture. The DQN model and the graph attention are the only trainable components. The language model is a frozen component. The solid lines indicate gradient flow.
    }
    \label{fig:kgdqn}
\end{figure*}

\subsection{Reward Function}
\label{subsec:rewardf}

Our policy model is trained using reinforcement learning.
The reward function must produce greater reward for sentences that are more likely to be found near a goal and less reward for sentences that are less likely to be found near a goal. 
For purposes of experimental simplicity, a goal is a sentence that contains a verb that is a member of a given VerbNet verb class (e.g., \texttt{admire}, which encapsulates verbs such as ``love''). 
Any VerbNet class can be chosen, though the choice affects goal achievement;
common verb classes result in high achievement regardless of model, and sparse verbs result in a poor reward signal. 
For example, ``discover'' has 43\% of the frequency of our most common verb class (``say''), and ``admire'' 25\%. We verified that goal rates are acceptable in our base corpus (Table \ref{Tab:Tcr}) before proceeding. 

{\em Reward shaping} refers to the pre-computation of rewards for a particular decision space based on heuristically computed approximations of true utility.
We use the same reward function adapted from \citet{tambwekar2019controllable}. 
Given an input prompt---a complete sentence describing an event in the story being generated---a continuation sentence is generated. 
Output is truncated at the first period or 20 tokens, whichever comes first.
The verb is extracted from the continuation and a reward $R(v)$ , which is made up of two components.
The first component computes the distance a sentence with a given verb class tends to be from the goal:
\begin{equation}
r_1(v) =\log \sum_{s \in S(v,g)}(\mathrm{len}(s) - \mathrm{dist}_s(v,g))\label{r1} 
\end{equation}
where $s\in S(v,g)$ is the set of all stories containing both the current verb class $v$ and the goal verb class $g$.
$\mathrm{len}(\cdot)$ is the number of sentences in the story and $\mathrm{dist}_s(\cdot)$ is the number of sentences in $s$ between that with $v$ and the sentence with $g$.
The second component computes the frequency that a verb class co-occurs in a story along with the goal:
\begin{equation}
r_2(v) =\log{\frac{\mathrm{count}(v,g)}{\mathrm{count}(v)}}\label{r2} 
\end{equation}
where $\mathrm{count}(v,g)$ is the number of stories containing the verb class $v$ and the goal verbclass $g$, whereas $\mathrm{count}(v)$ is the number of stories containing $v$.
The final reward is given as:
\begin{equation}
R(v) = \frac{1}{|verbs|} \times r_1(v) \times r_2(v) \label{R}
\end{equation}
For efficiency, the reward for all verbs are computed and stored in advance.
\subsection{Verb Clustering}

\citet{tambwekar2019controllable} observed that naively rewarding a story generator based on distance to a goal verb induces a model to skip to the end of a story in a single continuation.
To ensure that the model generates sequential sentences in the story and does not jump directly to the goal sentence, verbs are clustered 
depending on their reward value as calculated in equation \ref{R}.
Following \citet{tambwekar2019controllable}, we use the Jenks Natural Breaks optimization technique \cite{jenks1971error},
an off-the-shelf clustering algorithm appropriate for our 1-Dimensional non-uniform data.\footnote{We compute our Jenks clusters using JenksPy python library: \url{https://github.com/mthh/jenkspy}.} 
However, clusters computed with methods such as 1D $k$-means, or even quantiles or equal intervals would have still resulted in ordered clusters as described above.
The result is a set of ordered verb clusters estimating how ``close'' a verb class is to the goal verb class. These clusters provide further reward guidance, though applied differently depending on the specific reinforcement learning algorithm used.
During generation, the ``source'' is the most recent verb class's cluster, and the ``target'' is the next consecutive cluster in the direction of the goal cluster. 

\subsection{Knowledge Graph Guided Deep Q Network}
\label{subsec:kgdqn}

The policy model is trained to pick the best continuation from an un-tuned language model.

We formulate the problem of selecting a continuation sentence that moves the story closer to the goal as a 
Markov Decision Process (MDP) with the state being a knowledge graph that represents the current state of the story world.
A knowledge graph is a set of binary relation triples of the form $\langle subject, relation, object\rangle$.
This story world state representation explicitly captures the entities in a story and the relations between entities.
This constrains the state space to a discrete, but infinite, set of graphs.
It also allows the policy model to focus on elements of the story world that are likely to matter for the purposes of maintaining logical coherence and goal achievement. 

\subsubsection{Policy Model Architecture}
Our policy model is a variation of Deep $Q$ Networks (DQN).
A DQN takes a world state observation and predicts $Q(s,a_i)$ for state $s$ and each possible action $a_i$.
In the case of our system, a state is a knowledge graph and a continuation generated by GPT-2-sci-fi is an action.
Each action/continuation is embedded and concatenated to an embedding of the knowledge graph.
The knowledge graph is embedded we reuse GPT2-sci-fi word embeddings to embed graph nodes into vectors and then obtain a single vector representation using multi-headed graph attention \cite{veli2018graph}. 
The combined action and graph embedding vector is passed through a final fully connected layer to obtain predicted utility for that combination of graph and continuation. 

Knowledge graph triples are extracted from sentences using Stanford Open Information Extraction (OpenIE) \cite{2015angeli-openie}.  
OpenIE is a common and well-established method of extracting relation tuples, that has been demonstrated on related domains (cf.~\cite{kgdqnTextgames,ammanabrolu2020Graph}), and therefore a good proof of concept despite potential noise. 
The knowledge graph is updated with new triples after each sentence added to the story, thus creating an implicit transition function.
We train this policy network model by using the output of the base GPT2-Sci-Fi as if continuation sentences are actions and the knowledge graph as a discrete state, using the DQN algorithm \cite{mnih2015humanlevel}.
We refer to our full knowledge graph informed policy model as {\em KG-DQN} and is shown in Figure~\ref{fig:kgdqn}.

\subsection{Policy Model Training}

KG-DQN takes as input a knowledge graph and a potential action---a sentence generated by a frozen GPT2-sci-fi---and  outputs the expected utility $Q_\theta(G,a)$ of taking that action $a$ given the current state knowledge graph $G$. 
The standard DQN training loop~\cite{mnih2015humanlevel} populates an experience replay buffer, which collects up combinations of states, actions, and rewards.
The experience replay buffer is sampled and loss is computed relative to a target network, which is periodically replaced by a frozen version of the DQN being trained.
The full training loop is as follows: 
\begin{enumerate}
    \item Given a current query sentence $a_t$ and knowledge graph $G_t$ generate 25 potential candidates using GPT2-sci-fi.
    \item Clean candidates that do not contain complete sentences or verbs. Truncate the remaining candidates to form complete sentences. 
        \item Choose the next action $a_{t+1}$ $\epsilon$-greedily. The chosen action and updated knowledge graph will become inputs for the next time step.
        \item Using OpenIE, extract relevant knowledge triples and update knowledge graph to $G_{t+1}$. 
    \item Add $\langle G_t, a_{t+1}, G_{t+1}, reward, query\rangle$ tuple to our experience replay buffer. 
    \item Every 100 stories, sample our replay buffer and calculate $y_j = r_j$ for terminating actions, or $y_j = r_j + \gamma \textbf{max}_{a'} Q_\theta(G_{j+1}, a')$ for non-terminating ones.
    Perform gradient descent in $Q$ with loss $(y_j - Q_{\theta_t}(G_j, a_j))^2$ where $\theta_t$ is the frozen target network. 
    \item Every 300 stories, update the target network $\theta_t$. 
\end{enumerate}

During step 3 in the training loop, the agent can choose exploration with probability $\epsilon$ or exploitation. If exploitation is chosen 
we first tentatively prune out actions that do not lead from the current verb cluster to the next ``closest'' verb cluster to the goal. 
If that is impossible, we then fall back to picking the action/sentence according to the highest estimated reward from the set of all candidates. 
This is contrast to the standard approach of 
choosing the action with the maximum predicted utility from our policy. 
In this way we bias the model towards choosing cluster-optimal jumps and discourage behavior like jumping immediately to the goal (which would not create a coherent story).  

\section{Baselines and Additional Models}

All generation models are based on the GPT-2 117M model \cite{radford2019language} fine-tuned for story generation on NER-replaced science fiction data \cite{ammanabrolu2020story}, which we refer to as {\em GPT2-sci-fi}.
We experiment with five different models:
\begin{enumerate}
    \item {\bf GPT2-Sci-Fi}, the baseline model.
    \item {\bf GPT2-RS}, the base story generation model additionally fine-tuned with reward shaping based on the reward function in Section~\ref{subsec:rewardf}.
    This model is a near-literal update of \citet{tambwekar2019controllable} to work on GPT2 instead of a custom-trained seq2seq language model.
    Details of this model and its training are given below.
    \item {\bf KG-DQN}, our model, as described in Section~\ref{sec:methodology}.
    \item {\bf DQN}, Same as above but ablating/removing the knowledge graph representation.
    In this case, the state is the same as the action---the input sentence.
    This model has 36,138 trainable parameters compared to KG-DQN with 177,750 trainable parameters.
    \item {\bf KG-DQN-RS}, combining the KG-DQN policy with a frozen GPT2-RS fine-tuned network instead of GPT2-sci-fi as used in KG-DQN. 
\end{enumerate}

We did not include systems that use high-level plot guidance inputs (eg. ~\cite{fan2018hierarchical,peng2018towards,rashkin2020plotmachines}) as baselines because they accept additional guidance inputs that our technique does not have access to.
Nor do we include systems that are not goal-directed (eg., \cite{fan2019strategies,yao2019plan,peng2021inferring}) as baselines because they are not encumbered by the additional success criteria.
We also do not include infilling systems (eg. \cite{donahue2020ilm,ammanabrolu20automated,wang2020narrative}. 

We did not include the technique by \citet{tambwekar2019controllable} directly as a baseline because they used humans to transcribe event tuples into natural language sentences.
Our technique, however, is designed to generate natural language.
Without the translation, the event tuples are uninterpretable and with it there cannot be a fair comparison.
Unfortunately, the technique described by \citeauthor{tambwekar2019controllable} also cannot be directly applied to models that produce full-sentence outputs instead of event tuples.
One of the issues with fine-tuning GPT-2 on verb usage is that verbs that move the story closer to a goal ending may be rare.
Re-sampling (aka teacher-forcing) can be used when the vocabulary draws from event tuples.
However, swapping a verb without rewriting the entire continuation sentence can produce nonsensical results and was found not to improve goal-reaching behavior when applied to GPT-2.
Instead, our GPT2-RS baseline acts as an update to \citeauthor{tambwekar2019controllable} to account for neural language models that produce sentence outputs.
We update the technique with an alternative verb restriction approach to make sure that our model sees mostly verbs from our clusters during training. 
During training, for each given query, we generate twenty candidate sentences. We then we check if any of these sentences satisfies the following condition on the order and distance of source and target verbs' clusters:
\begin{enumerate}
\item The difference between the target verb cluster and the source verb cluster is one or zero. This sentence will get full reward based on Equation~\ref{R}.
\item The difference between the target verb and the source verb clusters is positive but higher than one. This means that the verbs are in the correct order but are further than they should be. Here, we discount the rewards by a factor of one over the difference between the clusters. 
\end{enumerate}
In all other cases, the model will give no rewards. This clustering ensures that the stories have reasonable lengths and reduce the model's bias to immediately produce sentences containing the goal verb.

As an initial step, the model is given a randomly sampled batch of the training sentence as query sentences; the objective is to fine-tune an underlying language model. 
The steps of training are: 
\begin{enumerate}
    \item The base language model generates a target sentence based on a query. The base model uses top-$k$ sampling in the generation. We choose $k=1000$ to produce better story results with less repetition in transformer models as stated in \cite{see2019massively}. If this target sentence has a verb with a positive difference then we move on to the next step. Otherwise, we sample a new query sentence from the training data and try again.

    \item The reward model then gives the reward amount this sentence should receive according to the criteria described above. 
    \item To fine-tune our base language model using the reward shaping function, we follow \cite{ziegler2019fine} approach in RL fine-tuning GPT-2 using the Proximal Policy Optimization algorithm (PPO2). We utilize the source sentences, target sentences, and assigned rewards to update the base model's (policy) loss. We utilize the Transformer Reinforcement Learning library\footnote{https://github.com/lvwerra/trl} which provides an implementation of PPO2 compatible with the Huggingface transformers library.\footnote{https://huggingface.co/transformers/index.html} 
    We train our models for 20 epochs each with batch size of 128, taking about 8 hours each.

\end{enumerate}
In time, the GPT2-RS training process fine-tunes the base language to prefer certain continuations based on how likely they are moving the story toward the given end-goal.

\section{Automated Evaluation}

\label{sec:eval}
To evaluate our models, we conduct an extensive set of automatic and human evaluations
in order to determine:
(1)~Does GPT2-RS-based RL fine-tuning with our described reward function work well when applied to GPT-2 for story generation? 
(2)~Is there a significant difference between using a policy gradient method (PPO) versus a value-based method like DQN? 
(3)~Does the explicit inclusion of knowledge graphs affect story generation performance?
Our automated evaluation metrics are:
\begin{enumerate}
    \item Goal Achievement Rate---the percentage of stories that produced a sentence with the target verb
    \item Average Perplexity scores
    \item 4-gram repetition~\cite{guan-etal-2020-knowledge}, measures the fraction of stories with at least one repeated 4-gram
    \item Average Story Length in number of sentences
\end{enumerate}
\subsection{Experimental setup}
\label{subsec:experimentalsetup}
We selected two verb class goals, \texttt{admire-31.2} and \texttt{discover-84}, 
chosen because they are sparse enough to not be generated by accident but not rare for stories. 
We then trained models for each goal.

The KG-DQN and DQN models were trained for 20 epochs each, and the best model every 5 epochs was taken for evaluation. We used a batch size of 256 and replay buffer of 800 using Adam. Our hyperparameters are discount factor = 0.99, learning rate = 0.001, $\epsilon$ = 0.1, with epsilon decaying following $\epsilon = (\epsilon - 0.01)/1000$ every training step. 

The GPT2-RS models were trained for 40 epochs each, and the best model every 10 epochs were taken for evaluation. We use similar settings to the ones  described in \cite{ziegler2019fine} with $KL_{\beta} = 0.1$, $KL_{target} = 6.0$, learning rate of $7.07\times 10^{-8}$ and 4 PPO epochs at each training epoch.

The KG-DQN-RS models simply combined the already trained components from KG-DQN and GPT2-RS respectively, with no further fine-tuning. 

We use our test set of 472 first sentences from the Sci-Fi stories dataset \cite{ammanabrolu2020story} as seeds to generate stories using our RL trained models and baseline models.  
All models could generate up to a maximum of 15 continuations to achieve the goal (for a total length of 16 sentences). 
Story generation terminated when the target goal was reached, the model failed to any more valid sentences, or 15 total sentences had been generated without meeting the goal condition. 
For generation, KG-DQN-RS, KG-DQN, and DQN were all run with a ``breadth'' of 25 candidate stories per step, generating from our test stories, with evaluation taking about 30 minutes per model. For Sci-Fi-GPT2 and GPT2-RS we ran them as generative models. 

\begin{table}
\centering
\footnotesize
\begin{tabular}{ |c|c|c|c|c|} 
\hline
{\bf Goal}& {\bf Model} & \makecell{{\bf Goal} \\ {\bf Rate}} & \makecell{{\bf Story} \\ {\bf Length}} & {\bf REP-4} \\ 
\hline
\multirow{4}{*}{\rotatebox[origin=c]{90}{\textit{admire}}} & Base Corpus & 32.42\% & 86.51 & 0.24 \\
& GPT2-Sci-Fi & 16.74\% & 15.01 & 0.3556  \\ 
& GPT2-RS & 41.95\% & 8.67 & 0.0339 \\ 
& DQN & 72.25\% & 4.54 & 0.2331 \\ 
& KG-DQN & 91.74\% & 4.84 & 0.1314 \\ 
& KG-DQN-RS & \textbf{94.70\%} & 5.11 & 0.0445\\
\hline 
\multirow{4}{*}{\rotatebox[origin=c]{90}{\textit{discover}}} & Base Corpus & 49.47\% & 86.51 & 0.24 \\
& GPT2-Sci-Fi & 18.86\% & 15.26 & 0.3369 \\ 
& GPT2-RS & 33.47\% & 9.71 & 0.0466 \\ 
& DQN & 96.19\% & 4.56 & 0.1123 \\ 
& KG-DQN & 98.09\% & 4.72 & 0.1123  \\ 
& KG-DQN-RS & \textbf{98.73\%} & 4.38 & 0.0466 \\
\hline
\end{tabular}
\caption{Results of Automated Experiments.}
\label{Tab:Tcr}
\bigskip
\begin{tabular}{ |c|c|c|} 
\hline
{\bf Goal} & {\bf Model} & {\bf Perplexity}\\ 
\hline
\multirow{2}{*}{\em admire}
& GPT2-Sci-Fi & 39.36 \\ 
& GPT2-RS & 40.034 \\ 
\hline 
\multirow{2}{*}{\em discover} 
& GPT2-Sci-Fi & 39.36\\ 
& GPT2-RS & 38.95\\ 
\hline
\end{tabular}
\caption{Perplexity Values}
\label{Tab:perplex}
\end{table}

\subsection{Automated Evaluation Results}

The results of our automated experiments can be found in Table \ref{Tab:Tcr}, and perplexity values in Table \ref{Tab:perplex}. 
We do not compute a perplexity for DQN models, as the generated tokens are from the base model, which is frozen. 
In order to best align results to compare models, in our model-generated stories, story length was only taken from stories that reached the goal. 
For the base-corpus, REP-4 was calculated over the first 16 sentences and not the full story. 

As can be seen from Table~\ref{Tab:Tcr}, in all our experiments KG-DQN-RS achieve the goal most often, slightly more often than KG-DQN. For both goals, GPT2-RS provided modest gains over baseline, but not as much as our DQN models. With both goal verbs, DQN models provide significant gains over baseline goal reaching behavior, with the knowledge graph augmented DQN outperforming the vanilla DQN, implying that knowledge graphs are important for goal reaching behavior (although in the \texttt{discover} goal the difference is less distinct). 
Also as can be seen from KG-DQN-RS, our DQN based approach is independent of the underlying language model used to generate candidate sentences. 

While the DQN models are able to achieve reductions in REP-4 score, the presence of GPT2-RS finetuning seems to account for the greatest reductions in repetition.

Although GPT2-RS fine-tuning alone was enough to bring goal-reaching over the GPT2-Sci-Fi baseline, perplexity is essentially unaffected, 

\begin{table*}[t]
\footnotesize
\begin{tabular}{|l|l|l|l|l|l|l|l|l|l|l|l|l|l|}
\hline
\multicolumn{2}{|l|}{\multirow{2}{*}{\bf Models}} &
  \multicolumn{2}{l|}{\bf Grammar} &
  \multicolumn{2}{l|}{\bf Avoids Repetition} &
  \multicolumn{2}{l|}{\bf Plausible Order} &
  \multicolumn{2}{l|}{\bf Local Causality} \\ \cline{3-10} 
\multicolumn{2}{|l|}{} &
  {\bf Win \%} &
  {\bf Lose \%} &
  {\bf Win \%} &
  {\bf Lose \%} &
  {\bf Win \%} &
  {\bf Lose \%} &
  {\bf Win \%} &
  {\bf Lose \%} \\ \hline
\multirow{5}{*}{\rotatebox{90}{\textit{admire}}} &
  KG-DQN-RS vs KG-DQN &
  \textbf{$24.81^{*}$} &
  61.29 &
  \textbf{$25.81^{*}$} &
  38.71 &
  \textbf{$9.68^{*}$} &
  45.16 &
  \textbf{$9.68^{*}$} &
  61.29 \\ \cline{2-10}
 &
  KG-DQN vs DQN &
  \cellcolor[HTML]{FFFC9E}29.03* &
  29.03 &
  \cellcolor[HTML]{9AFF99}\textbf{$35.48^{*}$} &
  22.58 &
  \cellcolor[HTML]{9AFF99}\textbf{$38.71^{*}$} &
  25.81 &
  \cellcolor[HTML]{9AFF99}\textbf{$45.16^{*}$} &
  22.58 \\ \cline{2-10}
 &
  GPT2-RS vs GPT2-Sci-Fi &
  \textbf{$6.45^{*}$} &
  38.71 &
  \cellcolor[HTML]{9AFF99}\textbf{$35.48^{*}$} &
  19.35 &
  \textbf{$19.35^{*}$} &
  32.26 &
  \textbf{$16.13^{*}$} &
  38.71 \\ \cline{2-10}
 &
  KG-DQN vs GPT2-Sci-Fi &
  \cellcolor[HTML]{9AFF99}\textbf{$58.06^{*}$} &
  12.90 &
  \cellcolor[HTML]{9AFF99}\textbf{$29.03^{*}$} &
  25.81 &
  \cellcolor[HTML]{9AFF99}\textbf{$61.29^{*}$} &
  19.35 &
  \cellcolor[HTML]{9AFF99}\textbf{$61.29^{*}$} &
  19.35 \\ \cline{2-10}
 &
  KG-DQN vs GPT2-RS &
  \cellcolor[HTML]{9AFF99}\textbf{$45.16^{*}$} &
  12.90 &
  \cellcolor[HTML]{9AFF99}\textbf{$48.39^{*}$} &
  16.13 &
  \cellcolor[HTML]{9AFF99}\textbf{$41.93^{*}$} &
  25.81 &
  \cellcolor[HTML]{9AFF99}\textbf{$38.71^{*}$} &
  16.13 \\ \hline
\multirow{5}{*}{\rotatebox{90}{\textit{discover}}} &
  KG-DQN-RS vs KG-DQN &
  \textbf{$12.5^{*}$} &
  50.0 &
  \textbf{$12.5^{*}$} &
  25.0 &
  \textbf{$15.63^{*}$} &
  56.25 &
  \textbf{$15.63^{*}$} &
  56.25 \\ \cline{2-10}
 &
  KG-DQN vs DQN &
  \cellcolor[HTML]{9AFF99}\textbf{$34.38^{*}$} &
  28.13 &
  \textbf{$25.0^{*}$} &
  31.25 &
  \cellcolor[HTML]{9AFF99}\textbf{$31.25^{*}$} &
  25.0 &
  \textbf{$28.13^{*}$} &
  43.75 \\ \cline{2-10}
 &
  GPT2-RS vs GPT2-Sci-Fi &
  \textbf{$16.13^{*}$} &
  51.62 &
  \cellcolor[HTML]{9AFF99}\textbf{$62.5^{*}$} &
  25.81 &
  \textbf{$22.58^{*}$} &
  48.39 &
  \textbf{$16.13^{*}$} &
  54.84 \\ \cline{2-10}
 &
  KG-DQN vs GPT2-Sci-Fi &
  \cellcolor[HTML]{9AFF99}\textbf{$71.88^{*}$} &
  9.38 &
  \cellcolor[HTML]{9AFF99}\textbf{$34.38^{*}$} &
  9.38 &
  \cellcolor[HTML]{9AFF99}\textbf{$45.16^{*}$} &
  22.58 &
  \cellcolor[HTML]{9AFF99}\textbf{$50.0^{*}$} &
  12.5 \\ \cline{2-10}
 &
  KG-DQN vs GPT2-RS &
  \cellcolor[HTML]{9AFF99}\textbf{$34.38^{*}$} &
  6.25 &
  \cellcolor[HTML]{9AFF99}\textbf{$38.71^{*}$} &
  16.13 &
  \cellcolor[HTML]{9AFF99}\textbf{$50.0^{*}$} &
  21.88 &
  \cellcolor[HTML]{9AFF99}\textbf{$46.88^{*}$} &
  25.0 \\ \hline
\end{tabular}
\caption{Human-participant pairwise evaluation results showing the percentage of participants who preferred the first model vs. the second. Each model was conditioned on the same eight first story sentences. * indicates significant results at $p<0.01$ confidence level using a Wilcoxon sign test on win-lose pairs. Green cells are wins of the first model, and yellow cells are ties. }
\label{tab:human}

\end{table*}

\section{Human Participant Evaluations}

\label{subsec:humanmetrics}
Human participant evaluations are believed to be the best practice in evaluating generated story quality. 
We asked human judges to compare pairs of stories generated by different models given the same input prompts.
Judges had to choose the better story (or equal) according to four criteria: 
\begin{compactitem}
\item \textbf{Grammar:} This story exhibits correct grammar.
\item  \textbf{Avoids Repetition:}  The story avoids repetition.
\item \textbf{Plausable Order:} This story’s events occur in a plausible order.
\item \textbf{Local Causality:} This story’s sentences make sense given sentences before and after them.
\end{compactitem}
These questions have been used in a number of other story generator evaluations~\cite{purdy2018predicting,tambwekar2019controllable,ammanabrolu2020story,ammanabrolu20automated,louisedgar,peng2021inferring}.
Plausible order and local causality questions are surrogates for {\em coherence}, which can be interpreted by human judges in different ways.

\subsection{Experimental Setup and Methodology}

As mentioned in Section \ref{prepdata}, our baseline GPT2-Sci-Fi was fine-tuned on NER-replaced data. This NER replacement makes story comprehension difficult for human subjects. 
We use pre-trained BERT to ``fill in the blanks'' using named entities from the original Sci-Fi dataset for the relevant category and taking the option with the best BERT-score \cite{havens2019fitbert}. We use these re-populated sentences for human evaluation. 

We chose a common set of seeds that were successes with all but GPT2-Sci-Fi and sampled a further subset of 80 seeds for human evaluation.

We recruited 64 participants on Prolific.\footnote{www.prolific.co}  
We presented each participant with five pairwise comparisons.  The story pairs are given to the participants in randomized order, ensuring that all pairs are seen equally by participants and that order does not impact the participants' answers. 

The average completion time for this task is 30 minutes, and the participants were compensated \$6  upon successful completion. To ensure high data quality, we added a checker question to ensure that the person reads and understands the tasks and a text field. In addition, every comparison asks the participants to explain their answers. 

\subsection{Human Participant Study Results}

We show the results of our pairwise comparison experiment in Table \ref{tab:human}. Our results show that participants preferred the stories generated GPT2-RS model over the baseline GPT2-sci-fi model in repetition avoidance for both goals.  
However, participants prefer the baseline in all other dimensions. 

The two-network KG-DQN model significantly ($p<0.01$) outperforms the GPT2-sci-fi baseline and GPT2-RS along all dimensions.
We note that participants often report ties; our analyses show that when participants report a difference, the difference significantly favors KG-DQN.
One reason for the high occurrence of ties is that it is easy for participants to escape tricky judgements by choosing to report a tie.

The DQN ablation (removing the knowledge graph from KG-DQN) shows a degradation of performance, strongly suggesting that the knowledge graph state representation is the factor in the reinforcement learning of the second network that plays the most important role.

When we add the reward shaped language model fine-tuning into KG-DQN to create KG-DQN-RS, we do not observe
much to improving the story generation except for repetition avoidance. 

\subsection{Qualitative Evaluation of Stories}
Even though our models do a great job generating stories that reach a specified goal, the generated stories' quality is fair with some noticeable issues. We notice that the longer the story gets, the more hallucinations make their way into the story diverging from the main topic even when the verbs are still going in the direction of the goal. Stories generated with the GPT2-RS model tend to be longer with many random character introductions, while the KG-DQN-RS, KG-DQN and DQN models are much more concise with a clear direction towards the goal. 
We see this is Table~\ref{Tab:Tcr}, where the DQN-based models tend to be shorter, achieving a goal verb after on average 4-5 events.
This is likely due to the verb clusters, which control how many ``hops'' the system has to take before arriving at the goal.
Table \ref{samplesadmire} and Table \ref{samplesdiscover} in Appendix A exemplify our models' output for a given prompt when trained on the goals: \texttt{Admire-31.2} and \texttt{Discover-84}.
 
\section{Conclusions}

Large-scale pre-trained language models are difficult to control, especially in the case of neural story generation in which one desires a story to end with a desired goal event.
While there has been some progress on making neural story generation models capable of progressing toward a given goal, those techniques do not transfer to large pre-trained language models. 
We show that reinforcement learning can be used to train policy models that are goal-aware and guide large language models.
These results provide a step toward story generation capabilities that benefit from the quality of language generation afforded by large pre-trained language models and also the ability to specify how a story ends, which up until now has not been achieved.

\bibliography{anthology,custom}
\bibliographystyle{acl_natbib}

\clearpage
\appendix
\section{Generated Story Samples}
In Table \ref{samplesadmire} and Table \ref{samplesdiscover} we show examples of some generated stories from our models. 

\begin{table*}[!ht]
\centering\begin{tabular}{|p{\textwidth}|}
 \hline
\textbf{KG-DQN-RS}

A Rogues scientist works in a Cavern, watched by a Rogues general and a guard. Alex and Blake finish working in ten forward. Alex makes an all-out attack on the team, killing all except Cameron. Blake dead, leaving them both shocked. Before Morgan decides to leave. B'le investigator kills Cameron. Alex turns around and heads back with Alex's investigator. \hl{Trusted}, Alex follows a guard into the murder of Robin. \\ \hline\hline
\textbf{KG-DQN}

A Rogues scientist works in a Cavern, watched by a Rogues general and a guard. Alex asks what will happen when the Manbot is removed. Hedgwick \hl{believes} that Blake will survive.  \\  \hline\hline
\textbf{DQN}

A Rogues scientist works in a Cavern, watched by a Rogues general and a guard. Alex's name is Blake. Cameron is watching the news. Alex does not \hl{believe} her and Cameron does not believe her. \\  \hline\hline
\textbf{GPT2-RS}

A Rogues scientist works in a Cavern, watched by a Rogues ran away, while Alex watches a happenance. I kept saying i kept from murdered like an fleas; you never turn life on me. Alex complains, but tries to tell her she's safe. J marketa pulls a gun on Alex. Morgan shoots at it, and then immediately throws it into the vapor. Blake notices the same thing from Cameron's office. Cameron postulates that to perfect the finisher after they work together, a normally impossible task. However Morgan base the idea one way or the other. Most especially Robin forcing Riley to put up with his section head problem. Meanwhile Cameron is having trouble depression. Alex and laranth are both having trouble \hl{believing} the story. \\  \hline\hline
\textbf{GPT-2 Sci-Fi} 

A Rogues scientist works in a Cavern, watched by a Rogues general and a guard. Alex watches them both die. Morgan and Blake look on. Ultimately killing each other. Bailey is taken to Cameron. Bailey is also saved in a holding Spacebarn, with the guard gone. Ellis and Morgan walk off together. Bailey is taken into custody, while Blake is taken into custody. Ellis is then taken to Hqs. Bailey is arrested, with the guard's body taken away. Ellis is then taken into custody. Blake is then taken into custody while the killer is taken into custody of Robin. Morgan is taken to a Safetech guard. Riley, under guards' knife. Jordan is then taken into custody and held for interrogation. Hqs, where Cameron is later held for questioning in a holding Spacebarn. Hqs, where Cameron is later taken to Morgan. \\  \hline\hline
\textbf{Goal:} Admire

\textbf{Prompt:} A ORGANIZATION0 scientist works in a LOCATION0, watched by a ORGANIZATION0 \\         \hline   
\end{tabular}
\caption{Story Output Examples from each model, for the goal Admire}
\label{samplesadmire}
\end{table*}

\begin{table*}
\begin{tabular}{|p{\textwidth}|}
  \hline
   \textbf{KG-DQN-RS:} 
   
   The tobacco company Hadara creates a form of super tobacco, which in turn is inhabited by a form of super tobacco beetle, which survive the cigarette processing and whose eggs are contained in the smoke of these killer cigarettes, presumably called brand Alex. Cameron, acknowledges the smoking man as being Alex. Morgan, the man confirms that he and his partner are actually doing the same thing. Bailey and Blake look down at the dead man, reply that they had better go. Alex \hl{seeing} Cameron dead but doors closing.  \\ \hline\hline
\textbf{KG-DQN:}

The tobacco company Hadara creates a form of super tobacco, which in turn is inhabited by a form of super tobacco beetle, which survive the cigarette processing and whose eggs are contained in the smoke of these killer cigarettes, presumably called brand Alex. Blake's research into the beetles also wasashes on the ground. Upon \hl{seeing} that Blake is not who she is, Blake goes to see Blake.   \\  \hline\hline
\textbf{DQN:}

The tobacco company Hadara creates a form of super tobacco, which in turn is inhabited by a form of super tobacco beetle, which survive the cigarette processing and whose eggs are contained in the smoke of these killer cigarettes, presumably called brand Alex. Cameron, in Millennia time, \hl{discovers} the truth.  \\  \hline\hline
\textbf{GPT2-RS:}

The tobacco company Hadara creates a form of super tobacco, which in turn is inhabited by a  Alex woman Blake runs across. Witnesses a b Bothhuman carrying a boy. Went into the autopsy site while he went walking. Cameron finds blood on the wounds and goes Invanova to investigate. Cameron notes red blood evacuate from the victims as well andhead stationed by Morgan. Shortly afterwards, they find Robin heading toward the Carkoon. Cameron says Robin has answered and she's coldly blamed Robin. However, Robin insists that she and she just have a disappearances. I don't know what i'm supposed to do. I'm not supposed to die this time, Blake informs Blake's. Pleased with that, Blake job outlook. K-7i tells them that the incident was simply a result of attacks. Cameron insists that something to do with it. Androgum brushes aside an idea that prove to be impossible. Down on Robin who drops a Grenade on Morgan, but it's too late. Morgan runs over to the fire escape where Morgan \hl{finds} a resources needed by those releases. \\  \hline\hline
\textbf{GPT-2 Sci-Fi:}

The tobacco company Hadara creates a form of super tobacco, which in turn is inhabited by a form of super tobacco beetle, which survive the cigarette processing and whose eggs are contained in the smoke of these killer cigarettes, presumably called brand Alex. Shortly thereafter, the beetle kills Blake, and they open fire in a room. Unfortunately, there is no response. They find a small explosion in the room, just as the beetle destroyed their factory. Cameron, alone with Jorjie, is captured by the insects. Hedril is able to escape. Blake manages to escape. Morgan is captured with Robin, Cameron's body cuffed. Riley and Jordan also find themselves both restrained. Riley and held. Jordan is taken into Jaime's chamber. Alex is taken into custody as a possible suspect. And told of the gun charges Riley brought to Alex's Homeworld's. Shelkonwa is brought to Riley, but then taken to Riley's Homeworld. Where she's questioned by an angry autopsy. Who demands to be sent before Alex's assistant, a state of mind. States that Alex is trying to see if Blake was really Blake a suspect.  \\  \hline\hline
 \textbf{Goal:} Discover
 
 \textbf{Prompt:} The tobacco company LOCATION0 creates a form of super tobacco, which in turn is inhabited by a \\         \hline   
\end{tabular}
\caption{Story Output Examples from each model, for the goal Discover}
\label{samplesdiscover}
\end{table*}

\section{Computing Infrastructure}
Our GPT2-RS models were trained and evaluated using Google Colab, on a Nvidia Tesla V100 GPU. KG-DQN and DQN were trained and evaluated locally on two machines: the first machine had 2 Nvidia GeForce GTX 1080 GPUs, the second had four Nvidia GeForce GTX 2080Ti GPUs.

\end{document}